\documentclass[runningheads]{llncs}

\usepackage{eccv}

\usepackage{eccvabbrv}

\usepackage{graphicx}
\usepackage{booktabs}

\usepackage[accsupp]{axessibility}  

\usepackage{multirow}
\usepackage{tikz}
\def\checkmark{\tikz\fill[scale=0.4](0,.35) -- (.25,0) -- (1,.7) -- (.25,.15) -- cycle;} 

\definecolor{cvprblue}{rgb}{0.21,0.49,0.74}
\definecolor{green1}{RGB}{34,139,34}
\usepackage{float}

\newtoggle{comments}
\toggletrue{comments}
\iftoggle{comments}{%
    \newcommand{\tarasha}[1]{{\leavevmode\color{magenta}[Tarasha: #1]}}
    \newcommand{\wesley}[1]{{\leavevmode\color{magenta}[Wesley: #1]}}
    \newcommand{\peiyun}[1]{{\leavevmode\color{red}[Peiyun: #1]}}
    \newcommand{\achal}[1]{{\leavevmode\color{orange}[Achal: #1]}}
    \newcommand{\david}[1]{{\leavevmode\color{purple}[David: #1]}}
    \newcommand{\deva}[1]{{\leavevmode\color{blue}[Deva: #1]}}
    
}{%
  \newcommand{\tarasha}[1]{}
  \newcommand{\wesley}[1]{}
  \newcommand{\peiyun}[1]{}
  \newcommand{\achal}[1]{}
  \newcommand{\david}[1]{}
  \newcommand{\deva}[1]{}
  
}

\def\maketitleappendix
   {
   \newpage
    \begin{center}
    {\Large\bfseries Appendix}
  \end{center}
   }

\usepackage[pagebackref,breaklinks,colorlinks]{hyperref}

\usepackage{orcidlink}

\begin{document}

\title{TAO-Amodal: A Benchmark for Tracking Any Object Amodally} 


\author{Cheng-Yen (Wesley) Hsieh\textsuperscript{1}
\and
Kaihua Chen\textsuperscript{1}
\and
Achal Dave\textsuperscript{2}
\and\\
Tarasha Khurana\textsuperscript{1}
\and
Deva Ramanan\textsuperscript{1}
}

\authorrunning{C-Y. et al.}
\institute{Carnegie Mellon University \and Toyota Research Institute\\
{\small \url{https://tao-amodal.github.io}}}

\maketitle

\begin{center}
        \centering
        \includegraphics[width=1.0\textwidth]{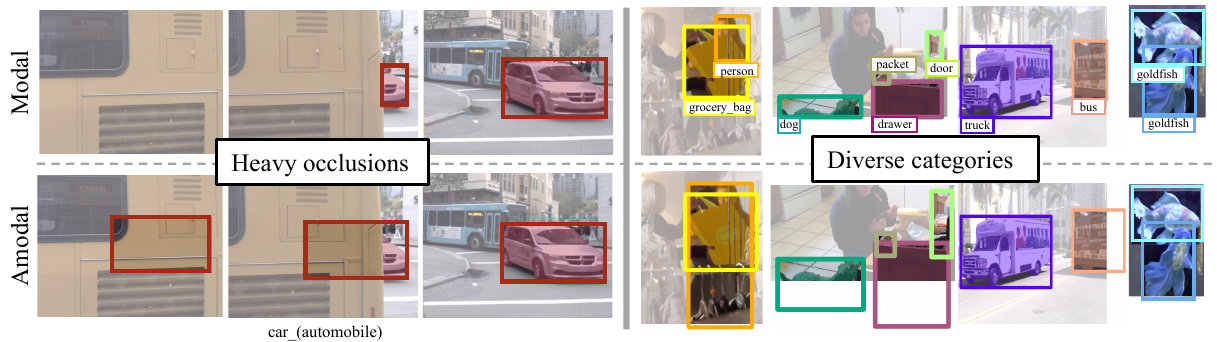}
        \captionof{figure}{\textbf{TAO-Amodal.} 
        We present TAO-Amodal, a dataset of amodal (bounding box) annotations for fully occluded and partially occluded (both within the image frame and out-of-frame) objects in videos from the TAO dataset~\cite{dave2020tao}. 
        Our dataset consists of 332k boxes that cover multiple occlusion scenarios across 2,907 videos with annotations for 833 object categories. TAO-Amodal aims at assessing the occlusion reasoning capabilities of current trackers for amodal tracking of any object.}
        \label{fig:representative}
\end{center}

\begin{abstract}
Amodal perception, the ability to comprehend complete object structures from partial visibility, is a fundamental skill, even for infants. Its significance extends to applications like autonomous driving, where a clear understanding of heavily occluded objects is essential. However, modern detection and tracking algorithms often overlook this critical capability, perhaps due to the prevalence of \textit{modal} annotations in most benchmarks.
To address the scarcity of amodal benchmarks, we introduce TAO-Amodal, featuring 833 diverse categories in thousands of video sequences.
Our dataset includes \textit{amodal} and modal bounding boxes for visible and partially or fully occluded objects, including those that are partially out of the camera frame.
We investigate the current lay of the land in both amodal tracking and detection by benchmarking state-of-the-art modal trackers and amodal segmentation methods. We find that existing methods, even when adapted for amodal tracking, struggle to detect and track objects under heavy occlusion. To mitigate this, we explore simple finetuning schemes that can increase the amodal tracking and detection metrics of occluded objects by 2.1\% and 3.3\%.

\end{abstract}

\begin{figure}[t]
    \centering
    \includegraphics[width=1\linewidth]{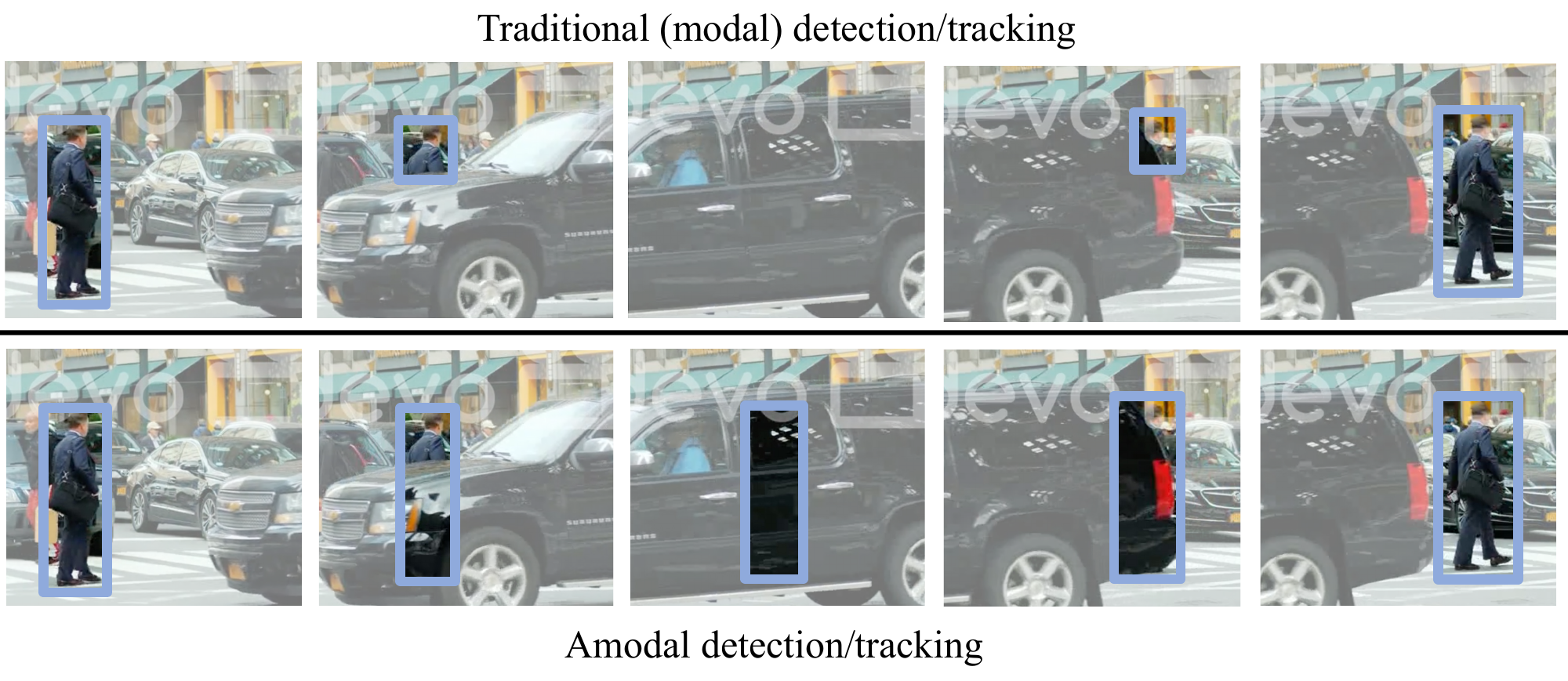}
    \caption{\textbf{Traditional modal perception (top) \textit{vs.} amodal perception (bottom).} Given a sequence of images, traditional detection and tracking algorithms concentrate on identifying visible segments of multiple objects within the scene. Consequently, they face challenges resulting in perculiar output such as vanishing bounding boxes or tiny box sizes under occlusion scenarios. Amodal perception advances beyond conventional approaches by inferring complete object boundaries, thereby predicting bounding boxes that extend to the full object extent, even when certain portions are occluded.}
    \label{fig:splash}
\end{figure}
\section{Introduction}
\label{sec:intro}

Machine perception, particularly in object detection and tracking, has focused primarily on reasoning about \textit{visible} or modal objects. This modal perception ignores parts of the three-dimensional world that are \textit{occluded} to the camera. However, amodal completion of objects in the real-world (e.g., seeing a setting sun but understanding it is whole) and their persistence over time (e.g., person walking behind a car in Fig.~\ref{fig:splash}) are fundamental capabilities that develop in humans in their early years~\cite{kavsek2004influence,InfantAmodalCompletion,InfantObjectPermanence}. In autonomous systems, this online amodal reasoning finds a direct application in downstream motion planning and navigation. Despite this, object detection and tracking stacks give little importance to partially or completely occluded objects; this becomes apparent in datasets that are only annotated modally~\cite{voigtlaender2019mots,openimages,gupta2019lvis,lin2014microsoft,everingham2010pascal,fan2019lasot,kristan2018sixth,yu2020bdd100k,dave2020tao,cioppa2022soccernet} but are still widely used and built upon by algorithms. These algorithms~\cite{ViTDet,QDTrack,GTR,AOA,meinhardt2022trackformer,maskdino,ssprl,TET,ren2015faster} in turn learn to perceive only modal objects.

To address this gap, we introduce a benchmark for large-scale amodal tracking, which requires estimating the full extent of objects through heavy and even complete occlusions. 
Our benchmark annotates 17,000 objects with amodal bounding boxes, along with human confidence estimates, from 833 classes in 2,907 videos.
While prior datasets focus on images or are limited to a small vocabulary of classes (Tab.~\ref{tab:stats}), our benchmark evaluates amodal tracking for hundreds of object classes.
Since objects can get occluded because of other objects in the scene, \textit{and} because of the limited field-of-view of cameras in casual captures, we define and address two kinds of occlusions -- in-frame, and out-of-frame.
As annotating amodal bounding boxes can be ambiguous and challenging, we design a new annotation protocol with detailed guidelines to improve human annotation.
Importantly, we base our benchmark on a large-vocabulary multi-object tracking dataset, TAO~\cite{dave2020tao}.
This choice allows us to pair our amodal box annotations with class labels, modal boxes, and precise modal mask annotations~\cite{athar2023burst} collected in prior work. 


Equipped with this data, we set out to evaluate the difficulty of amodal tracking.
We evaluate using standard metrics, including detection and tracking AP, and variants \cite{khurana2021detecting} that evaluate tracking specifically under partial and complete occlusions.
As expected, we find that standard trackers trained with modal annotations do not suffice for amodal tracking. 

To adapt existing modal trackers into amodal ones, we finetune them on TAO-Amodal. The closest line of work to amodal tracking is amodal segmentation \cite{zhan2023amodal, li2016amodal, qi2019amodal}. We benchmark recent amodal segmentation algorithms by running a Kalman-Filter based association during post-processing on their predictions. While this addresses the gap between modal and amodal tracking to some extent, the performance is far from good due to the challenging occlusion scenarios in TAO-Amodal. To mitigate this, we explore different but simple finetuning and data-augmentation strategies inspired by prior work \cite{li2016amodal,zhu2017semantic}. This lets us set a new baseline on the tasks of amodal detection and tracking. 


In summary, our contributions are as follows: (1) we annotate a large-scale dataset of amodal tracks for diverse objects, consisting of 17k objects spanning 833 categories, (2) we adapt evaluation metrics to handle amodal settings, and evaluate state-of-the-art trackers for our new task, and finally, (3) we investigate multiple finetuning and data-augmentation schemes as simple extensions to improve the existing modal tracking algorithms. 




\begin{table*}[t]
\caption{\textbf{Statistics of amodal datasets.} TAO-Amodal is proposed as an \textit{evaluation} benchmark for amodal tracking. We compare our dataset to prior image (first block), synthetic video (second block), and real video (last block) datasets. TAO-Amodal is notable for being {\em real-world} videos that span far {\em more categories} and far {\em more annotated frames} for evaluation. 
Track length is averaged over the dataset in seconds, while total length is the length of eval sequences in seconds.
We define heavy occlusion as objects with visibility below 10\%, and partial as between 10\%-80\%. Occluded tracks are those that have heavy or partial occlusions for more than 5 seconds. Out-of-frame (OoF) objects are ones that extend partially beyond the image boundary.}
    \label{tab:stats}
\resizebox{1.0\textwidth}{!}{
\begin{tabular}{lrrrrrrrrrrrrr}
 \toprule[1.5pt]
\multicolumn{1}{l}{}  & \multicolumn{4}{c}{\# Sequences} & & Track & Total & \multicolumn{3}{c}{\# Occluded Boxes} & \# Occluded   & Ann  \\
\multicolumn{1}{l}{} & Total & Test  & Val  & Train & Classes  & length & length & Partial & Heavy & OoF & tracks & fps \\ \midrule
COCO-Amodal~\cite{zhu2017semantic}  & 5000 &  1250   & 1250 & 2500  & - & - & - & 8.8k & 0.2k & 0 & -  & - \\ \midrule
Sail-VOS~\cite{hu2019sail}          & 201  & 0 & 41   & 160     & 162 & 14.14 & 3,359  & 559.5k & 704.8k & 0 & 7.9k & 8   \\
Sail-VOS-3D~\cite{hu2021sail}       & 202 & 0 & 41  & 161 & 24 & 13.10 & 2,808 & 295.0k & 387.5k & 0 & 5.0k  & 8                                                       \\ \midrule
NuScenes~\cite{caesar2019nuscenes}  & 1000 & 150 & 150 & \textbf{700} & 23  & 9.06 & 6,000 & 571.1k & \textbf{139.5k}  & \textbf{219k} & { \textbf{24.5k}}  & 20 \\
MOT17~\cite{milan2016mot16}         & 14 & 7 & 0 & 7 & 1  & 6.98 & 248 & 51.2k & 16.4k & 16k & 0.1k  & \textbf{30} \\
MOT20~\cite{dendorfer2020mot20}     & 8 & 4 & 0 & 4  & 1  & 20.55 & 178 & \textbf{729.4k} & 88.1k & 88k & 1.6k  & 25  \\ 
\begin{tabular}[c]{@{}c@{}}\textbf{TAO-Amodal}\end{tabular} & \textbf{2907} & \textbf{1419} & \textbf{988} & 500  & \textbf{833} & \textbf{22.24} & \textbf{88,605} & 158.2k & 35.1k & 139k & 9.6k & 1                
 \\
 \bottomrule
\end{tabular}
}
\end{table*}
\begin{figure}[h]
    \vspace{-7pt}
  \begin{minipage}[t]{0.48\columnwidth}
    \centering
    \includegraphics[width=\linewidth]{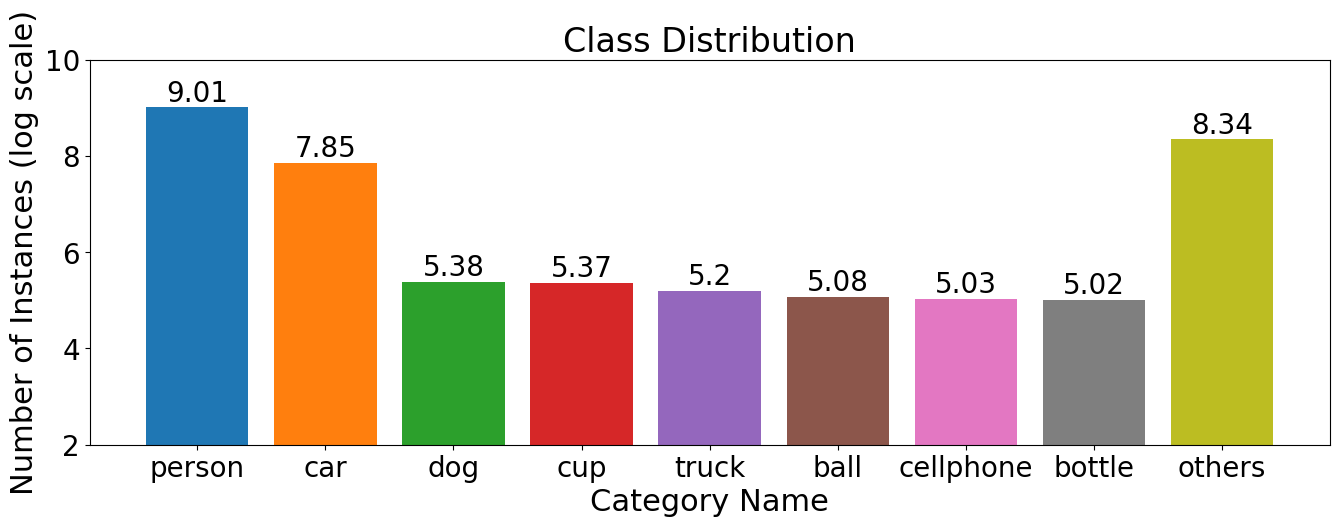} 
    \caption{\textbf{Class distribution.} We present counts of instances from top 8 most frequent categories and other categories, using a logarithmic scale.}
    \label{fig:class-distribution}
    \end{minipage}\hfill
  \begin{minipage}[t]{0.47\columnwidth}
    \centering
    \includegraphics[width=\linewidth]{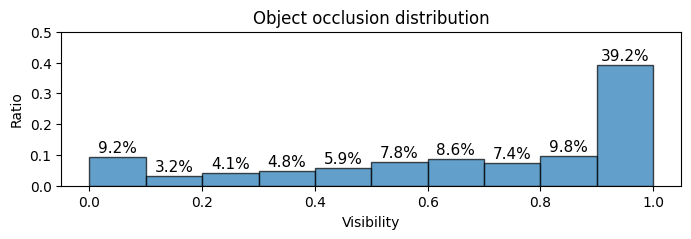} 
    \caption{\textbf{Object occlusion distribution.} We plot the distribution at a 10\% visibility span.}
    \label{fig:occlusion-distribution}
    \end{minipage}
\end{figure}

\section{Related work}
Amodal perception has been studied in the past by benchmarks and algorithms, in both the single-frame (detection) and multi-frame (detection and tracking) settings. Since amodal object annotations are hard to obtain due to the uncertainty in human annotations (c.f. prior work \cite{khurana2021detecting} on a human vision experiment), the community has depended heavily on synthetic datasets, or real-world datasets with few classes and limited diversity. We provide an overview of this prior wrok in the rest of this section. 

\subsection{Benchmarks}
\subsubsection{Real-world datasets.}
Amodal object annotations for real-world scenes are largely limited to the surveillance and self-driving domains.
MOT 15-20~\cite{leal2015motchallenge,milan2016mot16,dendorfer2020mot20} evaluate multi-object tracking on amodal person detections obtained from detectors trained on MOT annotations. However, these amodal annotations are automatically propagated via linear interpolation of annotations in frames where objects are visible. Additionally, the metrics used by MOT weigh all modal and amodal annotations equally. This largely ignores tracking performance on amodal objects, which form only a small fraction of all annotations. 

A number of multimodal (images and 3D LiDAR) datasets for autonomous driving have recently become popular. These include ArgoVerse (1.0 and 2.0)~\cite{chang2019argoverse,wilson2021argoverse}, Waymo~\cite{sun2020scalability}, nuScenes~\cite{caesar2019nuscenes} and KITTI~\cite{KITTI}. These datasets aim to focus on \textit{3D} tasks, and therefore use human annotators to label all objects in 3D to their full extent. In this setting, amodal annotations arise naturally due to the 3D nature of the data. These 3D boxes, when projected onto 2D images, would be useful for amodal perception;  unfortunately, these annotations cover only a small number of object classes. Another way to obtain amodal object annotations is in a multi-view setting. Datasets like CarFusion~\cite{reddy2018carfusion} and MMPTrack~\cite{han2023mmptrack} follow this data curation scheme, but, due to the cumbersome data collection process, they are limited to only a single or few categories.

In the single-frame setting, COCO-Amodal, Amodal KINS and NuImages~\cite{caesar2019nuscenes,zhu2017semantic,qi2019amodal} contain amodal annotations, but only cover the cases of partial occlusion: complete occlusions can only be recovered with temporal information, which is missing in image datasets. Moreover, COCO-Amodal~\cite{zhu2017semantic} and KINS~\cite{qi2019amodal} do not provide class labels, which makes it difficult to learn object priors for amodal completion, and to evaluate the accuracy of amodal tracking in the long tail.

\textbf{\\Synthetic datasets.}
An alternative approach to the above is use synthetic data generation pipelines to get amodal annotations. SAIL-VOS and SAILVOS-3D~\cite{hu2019sail,hu2021sail} are such datasets that exploit synthetic dataset curation and come with a number of different types of annotations (bounding boxes, object masks, object categories, their long-range tracks, and 3D meshes). Some of these even suit our case of detecting `out-of-frame' occlusions, where one could project 3D meshes onto the image plane. While the number of categories are slightly larger for these datasets (including others like ParallelDomain~\cite{paralleldomain} and DYCE~\cite{ehsani2018segan}), the sim-to-real transfer remains a challenge even for modal perception~\cite{chen2018domain,khodabandeh2019robust}.

\subsection{Algorithms}
Based off of some amodal datasets, there has been a growing interesting in developing algorithms suitable for amodal perception. 
Some methods aim to track objects with object permanence~\cite{khurana2021detecting,paralleldomain,tokmakov2022object,van2023tracking}. Previous work also segment objects amodally~\cite{li2016amodal,ling2020variational,zhan2023amodal,PCNet,orcnn}. Some approaches utilize prior-frame information~\cite{zhou2020tracking,cai2022memot,wu2021track,stearns2022spot,du2023strongsort,SORT,DEEPSORT,GTR}. For instance, GTR~\cite{GTR} employs a transformer-based architecture and uses trajectory queries to group bounding boxes into trajectories. We lean on similar approaches in this work, and devise a mechanism to generate occlusion cases in the flavor of the data augmentation used by GTR~\cite{GTR}, and show that this is essential to the goal of enabling amodal perception.

\begin{figure}[t]
    \centering
    \includegraphics[width=1.0\linewidth]{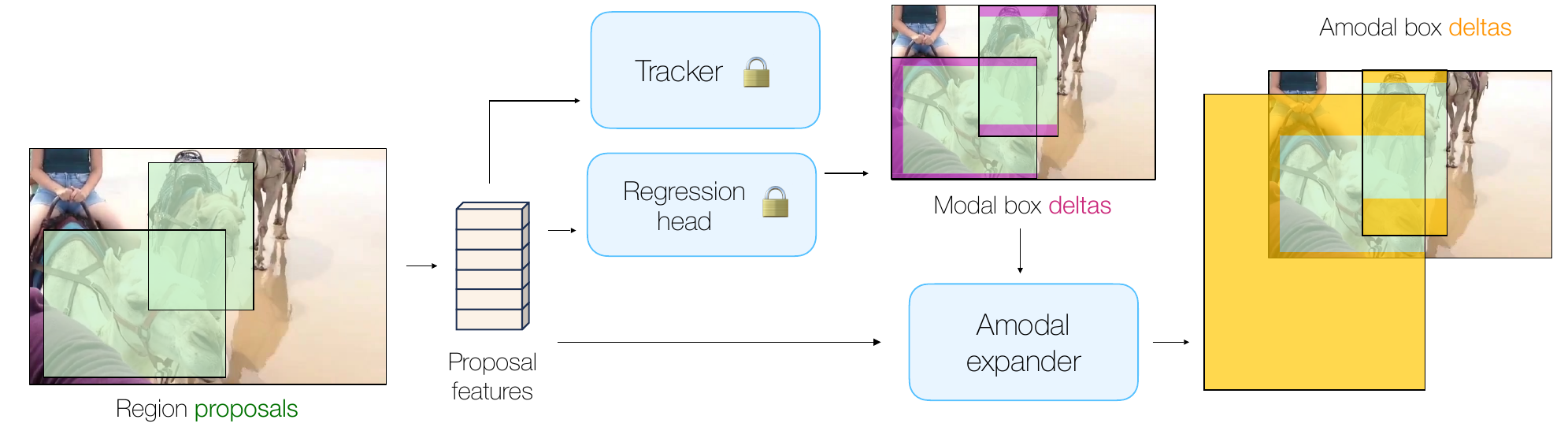}
    \caption{\textbf{ROI Head~\cite{girshick2015fast} with Amodal Expander.} 
    Amodal Expander serves as a plug-in fine-tuning scheme to ``amodalize" existing detectors or trackers with limited (amodal) training data.
    It operates by taking as input region proposal features and modal box predictions (often represented as a residual delta with respect the region proposal) and generates amodal box outputs (again represented as residual deltas). 
    We freeze all modules except the amodal expander during fine-tuning.}
    \label{fig:Amodal-Expander}
\end{figure}

\begin{figure}[t]
    \centering
    \includegraphics[width=0.95\linewidth]{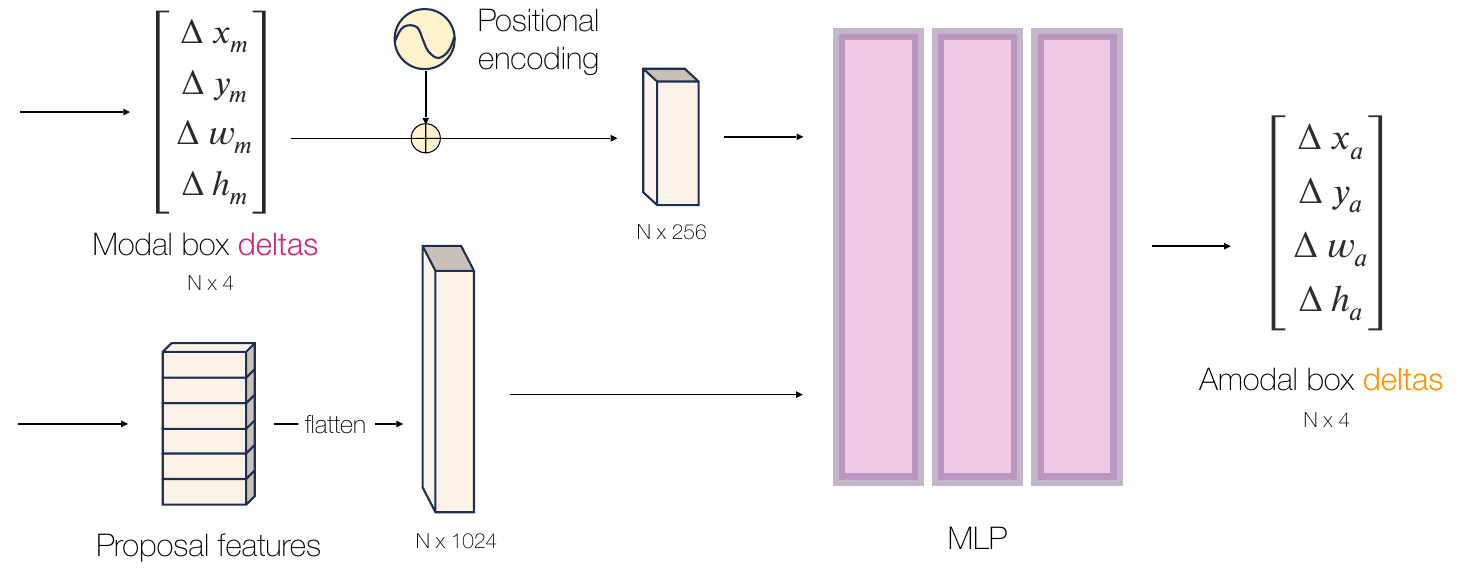}
    \caption{\textbf{Amodal Expander Architecture.} Given $N$ flattened proposal features and modal box (delta) predictions represented with 256-dim positional encodings~\cite{vaswani2017attention}, we predict amodal box (deltas) with a two-layer MLP (unless otherwise specified). Further architecture details are in Sec.~\ref{sec:exp_Amodal_Expander}.}
    \label{fig:Amodal-Expander-Detail}
\end{figure}
\begin{table*}[t]
    \centering
    \caption{\textbf{Amodal trackers on TAO-Amodal validation set.} We define metrics in Sec.~\ref{sec:benchmark}. The visibility range is indicated by the superscript to denote various levels of occlusion. We fine-tuned modal trackers on TAO-Amodal-train for 20k iterations. Detector~\cite{ViTDet} and amodal segmentation methods~\cite{PCNet,orcnn,aisformer} were evaluated using Kalman filter based association~\cite{SORT}. We evaluated models that predict COCO vocabulary~\cite{lin2014microsoft} using objects within COCO category. GTR is used as the basis for subsequent experiments, considering its performance in detection and tracking metrics. We run all trackers at 1 fps and average AP across categories with an IoU threshold of 0.5.}
    \label{tab:sota-fine-tune-AP50-1fps}
    \resizebox{0.8\textwidth}{!}{
    \begin{tabular}{lc@{\hskip 1em}ccccccc}
        \toprule[1.5pt] &
              & \multicolumn{5}{c}{Detection Metrics} & \multicolumn{2}{c}{Tracking Metrics} \\
\cmidrule(lr){3-7}\cmidrule(lr){8-9}
        Method & FT & AP$^{[0,0.1]}$ & AP$^{[0.1,0.8]}$ & AP$^{[0.8,1]}$& AP$^{\textrm{OoF}}$ & AP & AP & AP$^{\textrm{[0,0.8]}}$  \\
        \midrule
        PCNet~\cite{PCNet} &  & 0.48 & 7.15 & 21.43 & 8.69	& 15.59 & 5.80 & 3.91 \\
        QDTrack~\cite{QDTrack} & \checkmark & 0.35 & 8.03	& 21.82 & 8.05	 & 15.62 & 7.84 &4.03  \\
        TET~\cite{TET} & \checkmark & 0.24 & 5.77 & 14.98	& 4.87 & 10.86 & 4.84 & 3.44 \\
        ViTDet-B~\cite{ViTDet} & \checkmark & 0.77 & 12.57 &	34.33 &	14.18 & 	25.94 & 		7.66 &	4.38 \\
        ViTDet-L~\cite{ViTDet} & \checkmark & \textbf{1.25} & 15.06 & 38.16 & 15.84 & 29.04 & 9.70 & 5.90 \\
        ViTDet-H~\cite{ViTDet} & \checkmark & 1.13 & \textbf{15.80} & 	\textbf{40.09} & \textbf{16.97} & \textbf{30.20} & 9.72 & 5.63 \\
        GTR~\cite{GTR} & \checkmark & 0.77 & 14.62 & 38.17 & 15.31 & 29.24 & \textbf{16.07} &\textbf{ 9.28} \\
        \midrule
        \multicolumn{9}{c}{COCO category eval} \\
        \midrule
        ORCNN~\cite{orcnn}  & & 0.33 & 11.78 & 37.88 & 16.68 & 26.09 & 5.72 & 3.43 \\
        AmodalMRCNN~\cite{orcnn}  & & 0.46	& 14.74 & 42.65 & 18.35 & 29.58	& 7.57 & 4.47 \\
        AISFormer~\cite{aisformer}  & & 0.36 & 14.23 & 39.76 & 18.61 & 27.70 & 7.88	& 5.90 \\
        PCNet~\cite{PCNet}  & & \textbf{1.30} & \textbf{20.61} & \textbf{53.13} & \textbf{24.21} & \textbf{37.04} & \textbf{11.19} & \textbf{8.78}  \\
        
        \bottomrule[1.5pt]
    \end{tabular}
    }
\end{table*}

\begin{table*}[t]
    \centering
    \caption{\textbf{Exploring fine-tuning strategies on TAO-Amodal validation set.} We ablate different strategies for repurposing GTR for amodal tracking. Fine-tuning plug-in expander modestly outperforms fine-tuning all or part of the model. Combined with data augmentation, PasteNOcclude (PnO), for generating synthetic occlusions (as detailed in Sec~\ref{sec:PnO}), expander produces noticeable gains for partially occluded and out-of-frame objects. 
    All models (other than the baseline) were trained on TAO-Amodal training set for 20k iterations, while $^\dag$ denotes 45k iterations of training. 
    }
    \label{tab:amodal-expander-AP50}
    \resizebox{1.0\textwidth}{!}{
    \begin{tabular}{lllllllll}
        \toprule[1.5pt]
              &   \multicolumn{5}{c}{Detection Metrics} & \multicolumn{2}{c}{Tracking Metrics} \\
\cmidrule(lr){2-6}\cmidrule(lr){7-8}
        Method & AP$^{[0,0.1]}$ & AP$^{[0.1,0.8]}$ & AP$^{[0.8,1]}$ & AP$^{\textrm{OoF}}$ & AP & AP & AP$^{[0,0.8]}$ \\
        \midrule
        Baseline (GTR~\cite{GTR}) & 0.78 & 13.24 & 37.54 & 14.18 & 28.19 & 16.02 & 8.86 \\
        Fine-tune entire model & 0.52 & 10.36 & 24.08 & 10.34 & 17.93 & 7.70 & 3.93 \\
        FT entire model + PnO & 0.79 & 9.68 & 26.56 & 10.10 & 20.16 & 9.05 & 4.30 \\
        \begin{tabular}[c]{@{}c@{}}Fine-tune regression head\\ \& proposal network\end{tabular} & 0.79 & 10.57 & 27.91 & 11.37 & 21.42 & 9.04 &	4.53 \\
        Fine-tune regression head & 0.77 & 14.62 & 38.17 & 15.31 & 29.24 & 16.07 & 9.28 \\
        FT regression + PnO & 0.87 & 14.36 & 38.18 & 15.47 & 29.04 & 15.95 & 9.23 \\
        Amodal Expander & 0.67  & 16.29 & 37.11 & 17.39 & 29.50 & 16.10 & \textbf{10.44} (\textcolor{green1}{+1.58}) \\
        Amodal Expander + PnO & \textbf{0.80} (\textcolor{green1}{+0.02}) & 16.41 & 37.74 & 17.64 & 29.87 & \textbf{16.35} (\textcolor{green1}{+0.33}) & 10.13 \\
        Amodal Expander + PnO$^\dag$ & 0.77 & \textbf{16.53} (\textcolor{green1}{+3.29}) & \textbf{37.80} (\textcolor{green1}{+0.26}) & \textbf{17.65} (\textcolor{green1}{+3.47}) & \textbf{29.96} (\textcolor{green1}{+1.77}) & \textbf{16.35} (\textcolor{green1}{+0.33}) & 10.28 (\textcolor{green1}{+1.42}) \\
        \bottomrule[1.5pt]
    \end{tabular}
    }
\end{table*}

\section{Dataset Annotation}
\label{sec:annotation}

\subsubsection{Base dataset.} Existing datasets for modal perception are limited either in terms of their diversity, or the vocabulary of classes labelled. To this end, we build upon the modally annotated TAO dataset. It contains bounding box track annotations of 833 object categories at 1FPS spanning a total of 2,921 videos from 7 different data sources (AVA~\cite{gu2018ava}, Argoverse~\cite{chang2019argoverse}, Charades~\cite{sigurdsson2016hollywood}, HACS~\cite{zhao2019hacs}, LaSOT~\cite{fan2019lasot}, BDD100K~\cite{yu2020bdd100k}, YFCC100M~\cite{thomee2016yfcc100m}). Bootstrapping from this dataset allows us to add amodal box annotations to an already existing set of multimodal annotations in TAO -- i.e., object classes, modal bounding boxes and modal segmentation masks. TAO follows the single-frame detection datasets, such as LVIS and OpenImages~\cite{gupta2019lvis,openimages}, in adopting a federated annotation protocol for object tracking: i.e., not every object class is exhaustively annotated in every video. We refer the reader to~\cite{dave2020tao,gupta2019lvis} for details on federated annotation and evaluation setup, and focus here on our amodal annotation of objects in TAO. 

\textbf{\\Scope.} Since annotators can exhibit a large variation in annotating the precise shape of objects while they undergo partial or even complete occlusion, we annotate using bounding boxes instead of segmentation masks to mark the full extent of objects in the visible scene.
We define `in-frame' occlusions as those occurring from the presence of occluders (which may be other dynamic objects, or static scene elements), and `out-of-frame' occlusions as those resulting from objects leaving the camera field-of-view.
We do not label the extent of occlusion in cases where an object may be partially present \textit{behind} the camera (e.g., a person holding the camera who has their hands visible in the image). For labelling `out-of-frame' occlusions, we need to fix bounds for annotation on the image plane. We ask annotators to work within an \textit{annotation workspace} that extends to twice the image dimensions in consideration, with the image itself horizontally and vertically center-aligned in this workspace.
\textbf{\\\\Annotation Protocol.} Since object tracks in TAO are modal in nature, extending boxes to account for in-frame and out-of-frame occlusions requires (1) (in the case of partial occlusion) complementing TAO bounding boxes with amodal boxes, and (2) (in the case of complete occlusion) adding new boxes to object tracks for occluded frames.
Out of a total of 358,862 boxes in TAO, our annotators modify 266,902 (74.4\%) to account for partial occlusions.
Further, TAO-Amodal introduces an additional 23,449 bounding boxes for frames where objects were invisible and unlabeled in TAO.
These annotations follow the guidelines detailed in the appendix, covering a wide range of both in-frame and out-of-frame occlusion scenarios.
Importantly, we only consider occlusion cases where an object has appeared in the scene before.
We exclude occlusions where an object might be partially behind the camera or outside the annotation workspace defined above.
Within the strict purview of the guidelines, when an object's location cannot be discerned confidently by the annotators, annotators are instructed to mark an \texttt{is\_uncertain} flag.
From the 23,449 boxes for invisible objects, 20,218 (85.8\%) boxes are annotated confidently (i.e., without the uncertain flag), indicating that there is inherent uncertainty in localizing objects when they undergo heavy occlusions (matching observations from prior work~\cite{khurana2021detecting} which indicate human uncertainty of object location under occlusion).

Finally, equipped with both modal and amodal annotations for all objects, we add a visibility field to the TAO-Amodal annotations, using the overlap (intersection-over-union) between the modal and amodal boxes as a proxy.

\textbf{\\Quality Control.} We conduct two rounds of professional quality checks on TAO-Amodal annotations: all bounding box annotations are refined twice by annotators. Finally, the authors of this work conducted a manual quality check reviewing 349 tracks from 7 randomly sampled videos, and found only 2 (<1\%) tracks without an uncertainty flag to be erroneous. Both tracks were for objects with complete occlusions (visibility 0.0\%) in the video. 
Our analysis show that nearly all inspected tracks ($>99\%$) are accurate, indicating the high-quality of amodal tracking annotations in TAO-amodal.

\subsection{Dataset statistics}
We compare the statistics of TAO-Amodal to other amodal benchmarks in Table~\ref{tab:stats}. For NuScenes, which only categorizes object visibilities into four buckets, we use interpolation to estimate the number of boxes below visibility 0.1 and 0.8. A few amodal datasets are omitted from the table, either because they have been incorporated into TAO-Amodal~\cite{chang2019argoverse,yu2020bdd100k} or because these datasets lack quantified visibilities for categorizing different occlusion scenarios~\cite{cioppa2022soccernet,sun2022dancetrack}. TAO-Amodal covers annotations across an extensive 833 categories, which can be used to learn and evaluate object priors in a large-vocabulary setting. Furthermore, TAO-Amodal features a \textit{10$\times$} longer evaluation duration, ensuring a comprehensive evaluation. 

\subsection{Dataset Splits Design}
Following TAO~\cite{dave2020tao}, we propose TAO-Amodal primarily as an \textit{evaluation} benchmark. We choose to make the `validation' and `test' set larger to reliably benchmark trackers. 
We are not the first to do this: datasets in the single-object tracking~\cite{dave2020tao} community have similarly focused on evaluation. With the success of foundation models trained on internet data, high quality \textit{evaluation} benchmarks are more important than ever, as evidenced in the NLP community (\eg, MMLU~\cite{mmlu}).
Our training set is constructed in the spirit of instruction-tuning datasets: a small amount of data to \textit{align} models to perform amodal tracking.
We also find that increasing the size of the train set (by including test videos as train) does not significantly improve accuracy: adding $4\times$ more training data only increases AP$^{[0.1,0.8]}$ from +3.3\% to +3.7\%, further validating our choice to dedicate limited annotated data for evaluation. See appendix for details.


\begin{table}[t]
    \centering
    \caption{\textbf{Multi-frame-aware amodal baselines on TAO-Amodal validation set.}
    We explore extensions to include multi-frame signals for fine-tuned expander. Following~\cite{khurana2021detecting}, we use a Kalman filter to predict the positions of occluded objects, augmented by a monocular depth estimator to filter out spurious predictions. This leads to an increase in AP$^{[0, 0.1]}$. Further, we integrate multi-frame cross-attended Re-ID features, feeding them into the expander with concatenation. This boosts tracking and out-of-frame metrics.
    }
    \label{tab:temporal-amodal-expander-AP50}
    \resizebox{1.0\textwidth}{!}{
    \setlength\tabcolsep{3pt}
    \begin{tabular}{lllllllll}
        \toprule[1.5pt]
              &   \multicolumn{5}{c}{Detection Metrics} & \multicolumn{2}{c}{Tracking Metrics} \\
\cmidrule(lr){2-6}\cmidrule(lr){7-8}
        Method & AP$^{[0,0.1]}$ & AP$^{[0.1,0.8]}$ & AP$^{[0.8,1]}$ & AP$^{\textrm{OoF}}$ & AP & AP & AP$^{[0,0.8]}$ \\
        \midrule
        Baseline (GTR~\cite{GTR}) & 0.8 & 13.2 & 37.5 & 14.2 & 28.2 & 16.0 & 8.9 \\
        Amodal Expander & 0.8 & \textbf{16.4} (\textcolor{green1}{+3.2}) & \textbf{37.7} (\textcolor{green1}{+0.2}) & 17.6 & \textbf{29.9} (\textcolor{green1}{+1.7}) & 16.4 & 10.1 \\
        \hspace*{0.3cm} + Kalman filter & 1.8 & 15.8 & 36.3 & 16.4 & 29.0 & 16.0 &	10.1 \\
    \hspace*{0.3cm} + Depth~\cite{khurana2021detecting} & \textbf{2.0} (\textcolor{green1}{+1.2}) & 16.1 & 36.8 & 16.8 & 29.4 & 15.9 &	10.0 \\
        \begin{tabular}[c]{@{}l@{}}Amodal Expander\\ \hspace*{0.3cm} + Temporal Re-ID \\ \end{tabular} & 0.7 & 16.2 & 37.7 & \textbf{17.8} (\textcolor{green1}{+3.6}) & 29.8 & \textbf{17.1} (\textcolor{green1}{+1.1}) &	\textbf{11.0} (\textcolor{green1}{+2.1}) \\
        \bottomrule[1.5pt]
            \vspace{-25pt}
    \end{tabular}
    \vspace{-30pt}
    }
\end{table}

\section{Amodal tracking}

\subsubsection{Traditional and amodal tracking.}
Given a sequence of images $I^1, I^2, ..., I^t$, tracking approaches aim to output modal bounding boxes $b$, trajectory identifiers $\tau$, and class labels $s$ for objects across all frames. If an object is partially occluded, the box marks only the visible extent of the object, as illustrated in Fig.~\ref{fig:splash}. We focus here on amodal trackers, which similarly take as input a sequence of images, but, in addition to the modal tracker outputs, they generate amodal boxes $b_a$, which cover the full extent of partially / fully occluded objects.


In practice, training an amodal tracker end-to-end is infeasible due to the limited amount of amodal training data.
We focus instead on transforming a conventional tracker into an amodal one by leveraging its understanding of modal objects. To do this, we introduce a light-weight class-agnostic module $E$.

\begin{table*}[t]
    \centering
    \caption{\textbf{Evaluating the `people' category.} 
    We follow the conventions of Table~\ref{tab:amodal-expander-AP50} but evaluate performance only on the people category.
    Fine-tuned expander shows improvements over modal baseline, which can be observed in \cref{fig:video-qualitative}.
    We posit that this dramatic performance increase comes from the fact that people is the most common category.
    PasteNOcclude (PnO) leads to a slight drop for this category, which suggests that adding synthetic (occluded) examples is more helpful for less common categories.} 
    \label{tab:amodal-expander-people-AP50}
    \resizebox{1.0\textwidth}{!}{
    \begin{tabular}{lllllllll}
        \toprule
              &   \multicolumn{5}{c}{Detection Metrics} & \multicolumn{2}{c}{Tracking Metrics} \\
\cmidrule(lr){2-6}\cmidrule(lr){7-8}
        Method & AP$^{[0,0.1]}$ & AP$^{[0.1,0.8]}$ & AP$^{[0.8,1]}$ & AP$^{\textrm{OoF}}$ & Overall & Overall & AP$^{[0,0.8]}$ \\
        \midrule
        GTR~\cite{GTR} & 0.29 & 37.15 & 71.49 & 42.07 & 53.81 & 17.47 & 14.39 \\
        FT regression head & 0.41 & 49.32 & 78.93 & 53.26 & 61.36 & 20.44 & 18.74 \\
        Amodal Expander & 2.26 & 71.64 & 84.07 & 73.74 & 74.22 & \textbf{26.77 (\textcolor{green1}{+9.30})} & 28.94  \\
        Amodal Expander$^\dag$ & \textbf{2.46 (\textcolor{green1}{+2.17})} & \textbf{71.86 (\textcolor{green1}{+34.71})} & \textbf{84.21 (\textcolor{green1}{+12.72})} & \textbf{73.96 (\textcolor{green1}{+31.89})} & \textbf{74.34 (\textcolor{green1}{+20.53})} & 26.72 & \textbf{28.95 (\textcolor{green1}{+14.56})} \\
        Amodal Expander + PnO & 1.94 & 69.87 & 83.86 & 72.58 & 73.20 & 26.68 & 28.76 \\
        Amodal Expander + PnO$^\dag$ & 1.99 & 70.23 & 84.00 & 72.85 & 73.38 & 26.61 & 28.64 \\
        \bottomrule
    \end{tabular}
    }
\end{table*}
\subsection{Amodal expander}
We design an amodal expander $E$, which serves as a plug-in module to conventional trackers. For each object, the amodal expander takes as input the modal box $b$ and an embedding $f$ (which can be extracted from the tracker), and generates amodal bounding boxes $b_a$. 

\subsubsection{Predicting amodal boxes in a residual manner.}
The amodal expander operates as a refinement step, similar to the second stage of two-stage detectors~\cite{ren2015faster,he2017mask} and trackers~\cite{GTR}.
We introduce the amodal expander in the context of GTR~\cite{GTR}, although it can be applied to most standard modal trackers.
As illustrated in Fig.~\ref{fig:Amodal-Expander}, GTR produces modal boxes $b$ with corresponding object features $f$, and subsequently refines $b$ through a regression head $R$ by predicting a modal box delta $\Delta b$.
Our amodal expander takes as input the modal box delta $\Delta b$ and object feature $f$ as input, generating an amodal box delta.
This delta is then applied to the modal proposal $b$ to generate amodal boxes $b_a$, denoted as $E(\Delta b, f) + b \approx b_a$.
The training of the amodal expander follows the training of regression head~\cite{ren2015faster} by matching box proposals with a ground truth and applying regression loss.
We first match modal box predictions $b$ to a \textbf{modal} ground truth $b^*$. Then, we apply the regression loss, selected as smooth L1~\cite{girshick2015fast}, with the corresponding amodal ground truth $b^*_a$:
\begin{align}
    L(b, \Delta b, f) = L_{reg}(E(\Delta b, f) + b, b^*_a)
\end{align}

As shown in Tab.~\ref{tab:ablation-matching}, the matching strategy is crucial for training the expander.
Fig.~\ref{fig:Amodal-Expander-Detail} illustrates the amodal expander, and we provide implementation details in Sec.~\ref{sec:exp_Amodal_Expander}. We demonstrate the effectiveness of amodal expander in \cref{tab:amodal-expander-AP50,tab:amodal-expander-people-AP50}.


\begin{table}[t]
    \centering
    \caption{\textbf{Ablation: Region proposal matching strategy.} Given that modal trackers generate modal proposals, an improved strategy involves matching region proposals with modal ground truth (GT) while applying regression loss to amodal predictions against the amodal GT. Both expander models are trained with Paste-and-Occlude (PnO) on TAO-Amodal training set for 20k iterations.}
    \label{tab:ablation-matching}
    \begin{tabular}{lccccc}
        \toprule[1.5pt]
              &   \multicolumn{3}{c}{Detection AP} & \multicolumn{2}{c}{Tracking AP} \\
\cmidrule(lr){2-4}\cmidrule(lr){5-6}
        Matching & AP$^{[0.1,0.8]}$ & AP$^{\textrm{OoF}}$ & AP & AP & AP$^{[0.1,0.8]}$ \\
        \midrule
        Modal GT & 13.96 & 14.92 & 28.64 & \textbf{16.45} & 8.96 \\
        Amodal GT & \textbf{16.41} & \textbf{17.64} & \textbf{29.87} & 16.35 & \textbf{10.13} \\
        \bottomrule[1.5pt]
    \end{tabular}
\end{table}
\subsection{Synthesizing occlusion with Paste-and-Occlude (PnO)}
To simulate occlusion scenarios during training, we use a data augmentation technique inspired by~\cite{zhu2017semantic,ghiasi2021simple}, which we refer to as Paste-and-Occlude (PnO).
Paste-and-Occlude functions by pasting object segments into the original images to act as occludees.
The segment collection comprises 505k objects extracted from LVIS~\cite{gupta2019lvis} and COCO~\cite{lin2014microsoft} images using segmentation masks.
For each input image, we randomly select 1 to 7 segments from the collection and paste them at arbitrary locations, allowing for partial extension beyond the image boundary to replicate out-of-frame situations.
Subsequently, we incorporate the ground truth boxes of the pasted segments into the original ground truth masks. 
We find that PnO leads to improvements in detection across all occlusion scenarios, shown in Table~\ref{tab:amodal-expander-AP50}.
We posit that this synthetic strategy is particularly important for the long-tailed nature of TAO-amodal, unlike COCO-amodal, where a similar synthetic occlusion strategy leads to limited improvement~\cite{zhu2017semantic}.
We provide visual examples of synthetic occlusions in the appendix.

\label{sec:PnO}

\begin{figure*}[t]
    \centering
    \includegraphics[width=1.0\linewidth]{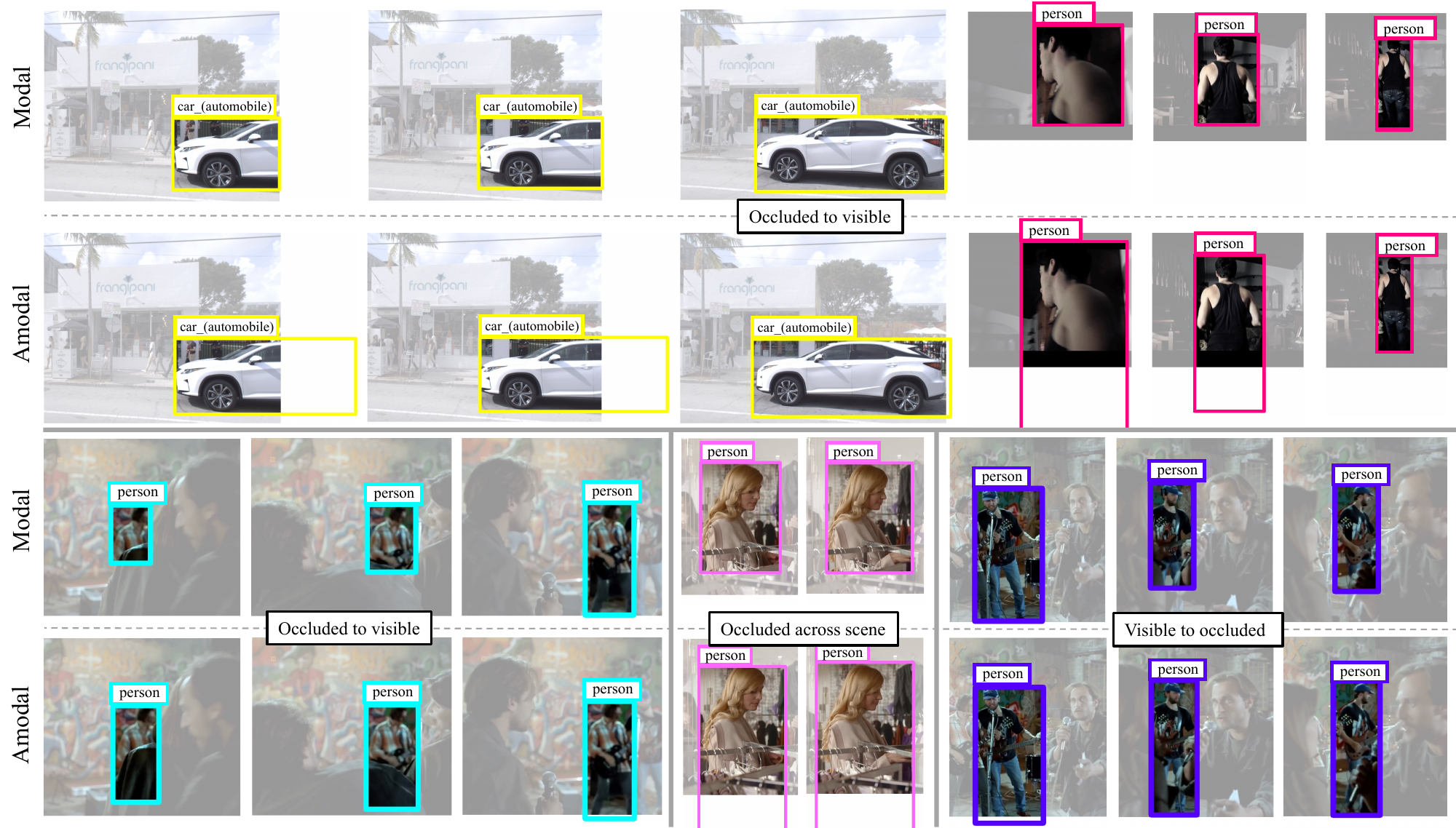}
    \captionof{figure}{\textbf{Qualitative results of Amodal Expander on TAO-Amodal val.} Trackers fine-tuned with expander produce both modal and amodal predictions. 
    The expander amodally complete objects that are occluded by objects in the scene (bottom-left) or objects that lie partially out of frame. We further verify that fine-tuned expander can amodally complete objects that were occluded in the past as well as objects that become occluded later.}
    \label{fig:video-qualitative}
\end{figure*}

\section{Empirical analysis}
In ~\cref{sec:benchmark}, we assess the challenges of amodal detection and tracking by evaluating a number of amodal trackers and segmentors.
Next, we investigate fine-tuning strategies and extensions for amodal baselines in \cref{sec:exp_Amodal_Expander}. 
We present implementation details and further ablations in the appendix.

\subsubsection{Evaluation Metrics.}
Using the estimated visibility attributes, we assess the tracking and detection capabilities of the model through variations of detection AP~\cite{lin2014microsoft} and Track-AP~\cite{dave2020tao}, representing the average precision across all categories at an IoU threshold of 0.5. 
We label objects with visibility less than 0.1 as heavily occluded, evaluated as AP\textsuperscript{[0.0, 0.1]}, where the superscript indicates the range of object visibility. If the visibility falls between 0.1 and 0.8, we categorize them as partially occluded, while those with visibility greater than 0.8 are considered non-occluded.
Objects that extend beyond the image boundary are referred to as out-of-frame (OoF) and evaluated with AP\textsuperscript{OoF}.
Additionally, we assess the model's performance on modal annotations with Modal AP.
In tracking, we evaluate highly or partially occluded tracks (Track-AP\textsuperscript{[0, 0.8]}), which are track with visibility at or below 0.8 for more than 5 frames (seconds).
We also evaluate performance on modal annotatiorns (Modal Track AP).
We provide a table of metric definitions in the appendix for quick reference. 

\subsection{Benchmarking state-of-the-art trackers}
\label{sec:benchmark}

\noindent \textbf{Evaluation of modal detectors and trackers.} We use three recent modal trackers, QDTrack \cite{QDTrack}, TET \cite{TET} and GTR \cite{GTR} and a detector, ViTDet \cite{ViTDet} for benchmarking. Every modal tracker is pre-trained on either TAO~\cite{dave2020tao} or LVIS~\cite{gupta2019lvis}, ensuring alignment of category vocabulary with our dataset. GTR~\cite{GTR} is trained on the combination of LVIS and COCO~\cite{lin2014microsoft} by generating synthetic videos using the training strategy in~\cite{zhou2020tracking}. QDTrack~\cite{QDTrack} and TET~\cite{TET} follow similar training procedures, pretraining detectors on LVIS and instance similarity heads on TAO for association. ViTDet~\cite{ViTDet} is trained on LVIS and combined with online SORT~\cite{SORT} tracker. We further adapt all models for amodal tracking by fine-tuning the regression head on TAO-Amodal training set for 20k iterations.



\textbf{\\Evaluation of off-the-shelf amodal segmentors.}
While we finetune modal algorithms for the amodal task, we note that these may not be architecturally optimized for amodal perception. To this end, we also evaluate state-of-the-art methods from the amodal segmentation line of work (note that mask prediction is more prevalent for amodal perception than box prediction). We benchmark ORCNN \cite{orcnn}, Amodal Mask-RCNN \cite{orcnn}, AISFormer \cite{aisformer} and PCNet \cite{PCNet}. ORCNN proposes a loss which brings occluders and ocludees spatially close. Amodal Mask-RCNN trains an additional amodal mask head on top of Mask-RCNN. AISFormer, also based on Mask-RCNN, uses transformer blocks in its amodal mask head to learn the spatial relations between visible and occluded objects. These methods only need an image as input and are trained on COCOA-cls~\cite{orcnn}. PCNet takes in modal masks of all objects in the scene as input, and recovers their relative ordering in the scene, before expanding modal masks into amodal ones. We use Detic \cite{detic} to get these modal masks. Finally, we run SORT \cite{SORT} on top of all boxes obtained from aforementioned methods and evaluate with proposed metrics only on COCO classes. PCNet shines likely because it only needs to \textit{expand} modal masks, which is a smaller lift than other baselines.


\textbf{\\How well do SOTA methods handle amodal perception?}
In \cref{tab:sota-fine-tune-AP50-1fps}, we see that both amodal segmentation baselines and fine-tuned modal trackers struggle in handling heavy occlusion and out-of-frame cases. To bridge the gap, we further explore different fine-tuning schemes and effects of data augmentation in \cref{tab:amodal-expander-AP50}, introduced in the next section. We report the performance of modal trackers on TAO-Amodal validation set as an ablation in the appendix.

\begin{figure*}[t]
    \centering
   \includegraphics[width=1.0\linewidth]{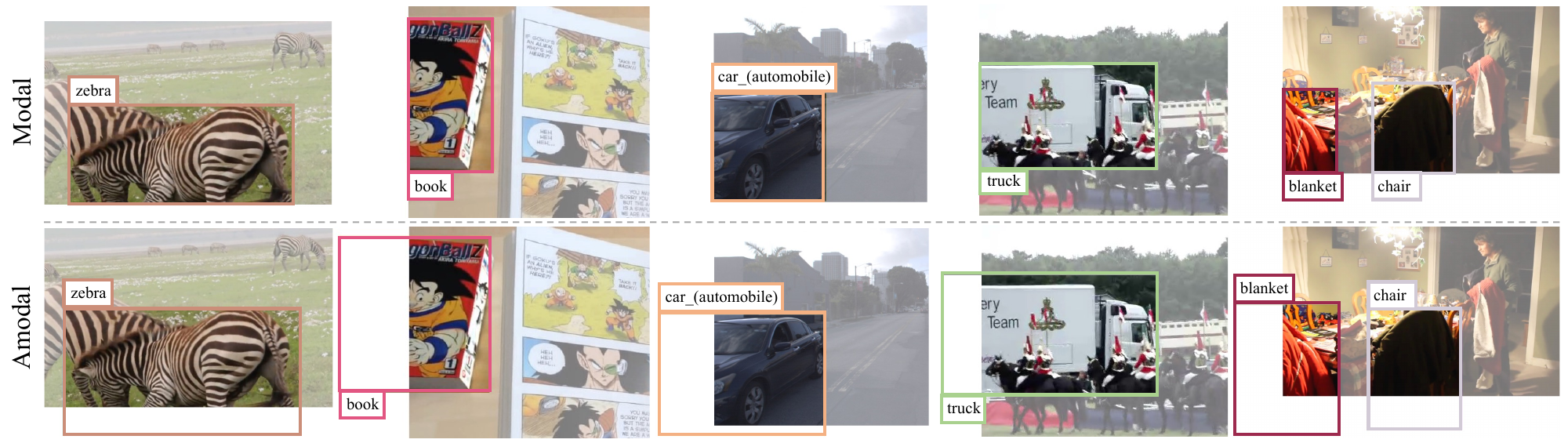}
    \captionof{figure}{\textbf{Qualitative results of Amodal Expander across diverse categories on TAO-Amodal val.} 
    Though we achieve the most impressive results for people, our Amodal Expander is effective across a diverse set of categories.}
    \label{fig:diverse-qualitative}
\end{figure*}

\subsection{Building amodal baselines with amodal expander}
\label{sec:exp_Amodal_Expander}
We illustrate amodal expander architecture in Fig.~\ref{fig:Amodal-Expander-Detail}. We build the expander on top of GTR~\cite{GTR} as this method shows reasonable performance in both detection and tracking aspects in Table~\ref{tab:sota-fine-tune-AP50-1fps}, likely due to its transformer-based association architecture that links identities over longer time periods with a sliding window of size 16. The hidden dimension of MLP is 256. We apply ReLU~\cite{relu} and dropout~\cite{srivastava2014dropout} with a probability of 0.2 to each layer except the last one. We train the amodal expander on the TAO-Amodal training set, along with PasteNOcclude (PnO) and augmentation used in GTR~\cite{GTR}. All the modules except the amodal expander are frozen during training. More ablation studies, hyperparameter details for training and PnO can be found in the appendix.

\textbf{\\Explore fine-tuning strategies for amodal perception.}
We explored several fine-tuning strategies including amodal expander on TAO-Amodal validation set as shown in~\cref{tab:amodal-expander-AP50}. Amodal expander trained with PnO for 45k iterations achieves 3.29\% and 3.47\% performance win under partially occluded (AP$^{[0.1,0.8]}$) and out-of-frame (AP\textsuperscript{OoF}) scenario. Fine-tuning entire model or solely the regression head and proposal network results in performance degradation. We posit that, with only 500 amodal training sequences, the models struggle to completely \textit{discard} modal knowledge. Fine-tuning box regression head is suboptimal when compared to amodal expander. Amodal expander further provides flexibility to adjust the architecture and select different input information, which are both important as shown in the ablation provided in appendix.

 \vspace{3mm}
\noindent{\bf Integrating multi-frame signals into amodal baselines.}
In Tab.~\ref{tab:temporal-amodal-expander-AP50}, we present two strategies for using temporal information within the amodal expander: 1) using a Kalman filter to forecast occluded object locations, with a monocular depth estimator to filter erroneous predictions, following~\cite{khurana2021detecting}, and 2) incorporating temporal Re-ID features. Note that (1) can associate single-frame detections, while also predicting \textit{new} boxes when an object is completely occluded. This significantly improves AP$^{[0, 0.1]}$.
 For (2), we take multi-frame Re-ID features and feed them into the amodal expander with channel concatenation. This helps handle out-of-frame occlusion while improving tracking metrics.

 \vspace{3mm}
\noindent{\bf Detecting people with amodal expander.}
In Table~\ref{tab:amodal-expander-people-AP50}, we study how well the expander baseline detects and tracks people, which serves as a crucial category in many autonomous driving and tracking benchmarks. Amodal expander obtains a significant improvement compared to the modal baseline, particularly on AP\textsuperscript{[0.1, 0.8]} and AP\textsuperscript{OOF}. Tracking on highly or partially occluded people (Track-AP\textsuperscript{[0.0,0.8]}) also increases by 14.6\%. This shows that one can obtain an effective amodal people tracker that could also track objects of diverse category vocabulary with our dataset using a simple fine-tuning scheme.

 \vspace{3mm}
\noindent{\bf Importance of proposal matching strategies.}
To apply regression loss, training a box prediction head requires matching each region proposal to a ground truth box. A naive strategy is to directly match the region proposals to the amodal ground truth box. However, direct matching with amodal boxes leads to suboptimal results as shown in~\cref{tab:ablation-matching}. As standard trackers generate modal region proposals, the model faced challenges in aligning proposals with the accurate ground truth due to a low Intersection over Union (IoU) between modal proposals and amodal ground truth. Matching proposals with modal boxes and applying regression loss using amodal ground truth yield better results.

\section{Discussion}
In this work, we focus on amodal perception of real-world objects. We draw inspiration from cognitive functions of amodal completion and object permanence in humans, that develop at an early age. Despite this, advancements in perception stacks like object detection and tracking, do not make amodal understanding central. We bring focus to three aspects/stages of building amodal perception stacks. First, we contribute a benchmark that annotates 833 categories of objects amodally in unconstrained indoor and outdoor settings, under partial and complete occlusion. Second, we contribute a benchmarking protocol in the form of metrics that evaluate detection and tracking specifically for the cases of partial or complete occlusions. Our key finding here is that existing algorithms struggle on these metrics. To address the observed limitations in amodal detection and tracking performance, and as a third contribution, we investigate data augmentation and fine-tuning schemes to boost existing tracking algorithms.
We hope our empirical evaluation provides a foundation for improving amodal perception.

%
%
\bibliographystyle{splncs04}
\bibliography{egbib}

\clearpage
\maketitleappendix
\begin{figure}
    \centering
    \includegraphics[width=1.0\linewidth]{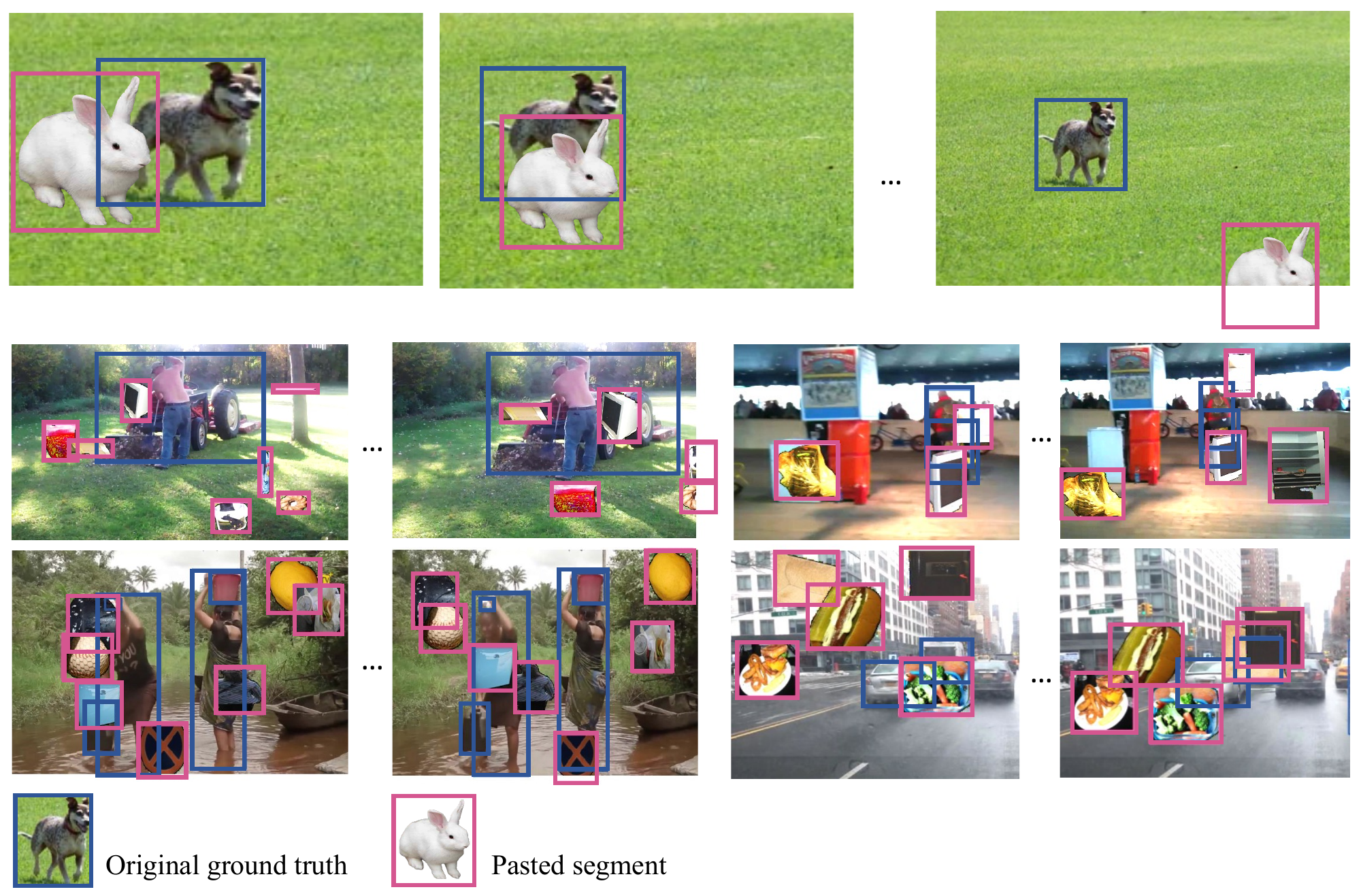}
    \caption{\textbf{Synthetic occlusions with PasteNOcclude (PnO).} PnO allows us to manually simulate occlusion scenarios and out-of-frame scenarios. We randomly choose 1 to 7 segments from a collection sourced from LVIS~\cite{gupta2019lvis} and COCO~\cite{lin2014microsoft} for pasting. For each inserted segment, we randomly determine the object's size and position in the first and last frames. The size and location of the segment in intermediate frames are then generated through linear interpolation.}
    \label{fig:PnO}
\end{figure}

\begin{table*}[t]
    \centering
    \caption{\textbf{Off-the-shelf trackers on TAO-Amodal validation set.} Off-the-shelf trackers were either trained on TAO~\cite{dave2020tao} or on synthetic videos~\cite{GTR} generated using LVIS images~\cite{gupta2019lvis}, with categories aligned with our dataset. While certain trackers can detect non-occluded objects well (over 35\% AP), objects that are highly occluded, partially occluded, and out-of-frame remain challenging, highlighting the difference between modal and amodal tracking. We run all existing trackers at 1 fps and average AP across all categories with an IoU threshold of 0.5. }
    \label{tab:Off-the-shelf-AP50-1fps}
    \resizebox{1.0\textwidth}{!}{
    \begin{tabular}{lccccccccc}
        \toprule[1.5pt]
              &   \multicolumn{6}{c}{Detection Metrics} & \multicolumn{3}{c}{Tracking Metrics} \\
\cmidrule(lr){2-7}\cmidrule(lr){8-10}
        Method & AP$^{[0,0.1]}$ & AP$^{[0.1,0.8]}$ & AP$^{[0.8,1]}$& AP$^{\textrm{OoF}}$ & Modal AP & AP & AP & AP$^{\textrm{[0,0.8]}}$ & Modal AP \\
        \midrule
        QDTrack~\cite{QDTrack} & 0.39 & 7.79 & 21.70 & 7.88 & 20.07 & 15.47 & 7.84 & 4.03 & 11.36 \\
        TET~\cite{TET} & 0.70 & 8.89 & 29.96 & 8.66 & 29.42 & 22.04 & 4.72 & 3.32 & 7.7 \\
        AOA~\cite{AOA} & 0.56 & 6.32 & 24.14 & 6.53 & 23.27 & 17.76 & 13.63 & 6.63 & 21.18 \\
        Detic + SORT~\cite{detic,SORT} & 0.38 & 6.68 & 21.31 & 8.09	& 18.84 & 15.32  & 6.18 & 3.81 & 8.16 \\
        ViTDet-B + SORT~\cite{ViTDet,SORT} & 0.77 & 11.40 & 34.03 & 12.98 & 32.67 & 25.15 & 6.95 & 4.10 & 11.57 \\
        ViTDet-L + SORT~\cite{ViTDet,SORT} & \textbf{1.18} & 13.75 & 37.41 & 14.70 & 36.65 & 28.05 & 8.19 & 5.14 & 13.73 \\
        ViTDet-H + SORT~\cite{ViTDet,SORT} & 1.03 & \textbf{14.54} & \textbf{39.71} & \textbf{16.53} & \textbf{38.05} & \textbf{29.56} & 8.94 & 5.76 & 14.55 \\
        GTR~\cite{GTR} & 0.78 & 13.24 & 37.54 & 14.18 & 36.08 & 28.19 & \textbf{16.02} & \textbf{8.86} & \textbf{22.50} \\
        \bottomrule[1.5pt]
    \end{tabular}
    }
\end{table*}

\begin{table*}[t]
    \centering
    \caption{\textbf{Off-the-shelf trackers on TAO-Amodal validation with higher IoU thresholds.} The definitions of our evaluation metrics can be found in Table~\ref{tab:Eval-Metrics}. The AP numbers are averaged over 10 IoU values from 0.5 to 0.95 with a 0.05 step, denoted as AP$_{0.5:0.95}$. We observed a similar performance trend as results evaluated with an IoU threshold 0.5. We run all trackers at 1 fps.}
    \label{tab:Off-the-shelf-AP-1fps}
    \resizebox{1.0\textwidth}{!}{
    \begin{tabular}{lccccccccc}
        \toprule[1.5pt]
              &   \multicolumn{6}{c}{Detection AP$_{0.5:0.95}$} & \multicolumn{2}{c}{Tracking AP$_{0.5:0.95}$} \\
\cmidrule(lr){2-7}\cmidrule(lr){8-9}
        Method & AP$^{[0,0.1]}$ & AP$^{[0.1,0.8]}$ & AP$^{[0.8,1]}$& AP$^{\textrm{OoF}}$ & Modal AP & AP & AP & AP$^{\textrm{[0,0.8]}}$ \\
        \midrule
        QDTrack~\cite{QDTrack} & 0.12 & 2.29 & 13.03 & 2.90 & 12.64 & 8.53 & 3.36 & 1.52 \\
        TET~\cite{TET} & 0.21 & 2.71 & 17.27 & 3.14 & 17.58 & 11.80 & 1.99 & 1.14 \\
        AOA~\cite{AOA} & 0.26 & 1.87 & 15.98 & 2.84 & 16.36 & 10.52 & 6.59 & 2.07 \\
        ViTDet-B + SORT~\cite{ViTDet,SORT} & 0.33 & 3.41 & 19.67 & 5.02 & 19.83 & 13.39 & 3.03	& 1.40 \\
        ViTDet-L + SORT~\cite{ViTDet,SORT} & \textbf{0.43} & 4.14 & 22.08 & 5.81 & 22.65 & 15.35 & 4.16	& 1.84 \\
        ViTDet-H + SORT~\cite{ViTDet,SORT} & 0.36 & 4.38 & 23.62 & \textbf{6.67} & 23.89 & 16.21 & 4.24	& 1.94 \\
        GTR~\cite{GTR} & 0.24 & \textbf{4.60} & \textbf{26.01} & 6.62 & \textbf{26.83} & \textbf{18.07} & \textbf{7.52} & \textbf{3.05} \\
        \bottomrule[1.5pt]
    \end{tabular}
    }
\end{table*}

\begin{table*}[t]
    \centering
    \caption{\textbf{Off-the-shelf trackers on TAO-Amodal validation set running at 5 fps.} ViTDet~\cite{ViTDet} achieves a performance gain by running at a higher fps as SORT~\cite{SORT} leverages its capability to estimate the new location based on the location in previous frames. AP numbers are averaged across all categories at an IoU threshold 0.5.}
    \label{tab:Off-the-shelf-AP50-5fps}
    \resizebox{1.0\textwidth}{!}{
    \begin{tabular}{lccccccccc}
        \toprule[1.5pt]
              &   \multicolumn{6}{c}{Detection AP} & \multicolumn{3}{c}{Tracking AP} \\
\cmidrule(lr){2-7}\cmidrule(lr){8-10}
        Method & AP$^{[0,0.1]}$ & AP$^{[0.1,0.8]}$ & AP$^{[0.8,1]}$& AP$^{\textrm{OoF}}$ & Modal AP & AP & AP & AP$^{\textrm{[0,0.8]}}$ & Modal AP \\
        \midrule
        QDTrack~\cite{QDTrack} & 0.42 & 7.59 & 21.53 & 7.78 & 19.98 & 15.42 & 6.63 & 2.72 & 10.34 \\
        TET~\cite{TET} & 0.24 & 5.39 & 14.56 & 4.73	& 29.42 & 10.51 & 3.52 & 2.21 & 5.56 \\
        AOA~\cite{AOA} & 0.56 & 6.29 & 24.35 & 6.77	& 23.51 & 17.85 & 12.82 & 5.53 & 20.67 \\
        ViTDet-B + SORT~\cite{ViTDet,SORT} & 1.00 & 13.38 & 37.98 & 14.78 & 37.08 & 28.32 & 10.09 & 4.40 & 16.93 \\
        ViTDet-L + SORT~\cite{ViTDet,SORT} & \textbf{1.32} & 16.38 & 43.30 & 17.16 & 42.31 & 32.08 & 11.75 & 5.53 & 19.22 \\
        ViTDet-H + SORT~\cite{ViTDet,SORT} & 1.06 & \textbf{17.24} & \textbf{45.18} & \textbf{18.58} & \textbf{44.02} & \textbf{33.53} &	13.16 & 5.87 & \textbf{21.39} \\
        GTR~\cite{GTR} & 0.57 & 12.45 & 35.89 & 13.63 & 34.92 & 27.28 & \textbf{13.70} & \textbf{7.02} & 20.09 \\
        \bottomrule[1.5pt]
    \end{tabular}
    }
\end{table*}

\begin{table}[t]
    \centering
    \caption{\textbf{Scaling up training data for amodal expander.} All fine-tuning is done on a set of 1,928 videos, vs. 500 in the main paper.}
    \label{tab:scale-amodal-expander-AP50}
    \resizebox{1.0\textwidth}{!}{
    \setlength\tabcolsep{3pt}
    \begin{tabular}{lllllllll}
        \toprule[1.5pt]
              &   \multicolumn{5}{c}{Detection Metrics} & \multicolumn{2}{c}{Tracking Metrics} \\
\cmidrule(lr){2-6}\cmidrule(lr){7-8}
        Method & AP$^{[0,0.1]}$ & AP$^{[0.1,0.8]}$ & AP$^{[0.8,1]}$ & AP$^{\textrm{OoF}}$ & AP & AP & AP$^{[0,0.8]}$ \\
        \midrule
        Baseline (GTR [59]) & 0.8 & 13.2 & 37.5 & 14.2 & 28.2 & 16.0 & 8.9 \\
        \midrule
        Fine-tune entire model & \textbf{1.1} & 12.7  & 29.1 & 12.4 & 22.5 & 9.7 & 6.2  \\
        Fine-tune regression head & 0.9 & 14.4 & \textbf{38.0} & 15.4 & 29.1 & \textbf{16.9} & 9.5 \\
        Amodal Expander & 0.8  & \textbf{16.9} (\textcolor{green1}{+3.7}) & 37.7 & \textbf{17.9} (\textcolor{green1}{+3.7}) & \textbf{30.0} (\textcolor{green1}{+1.8}) & 16.5 & 10.7 \\
        Amodal Expander + PnO & 0.7 & 16.5 & 37.8 & \textbf{17.9} (\textcolor{green1}{+3.7}) & \textbf{30.0} (\textcolor{green1}{+1.8}) & 16.5 & \textbf{10.8} (\textcolor{green1}{+1.9}) \\
        \bottomrule[1.5pt]
        \vspace{-25pt}
    \end{tabular}
    }
\end{table}

\begin{table}[t]
    \centering
    \caption{\textbf{Input to Amodal Expander.} Modal box (deltas) $\Delta b$, output by the regression head as shown in ~\cref{fig:Amodal-Expander}, contains information about the exact location of modal box predictions. Object features $f$ are embedded with visual appearance information of the modal proposals. We found that both information are important in amodally inferring the object's shape. All models were trained on TAO-Amodal training set with PasteNOcclude (PnO) for 20k iterations.}
    \label{tab:ablation-input}
    \begin{tabular}{lccccc}
        \toprule[1.5pt]
              &   \multicolumn{3}{c}{Detection AP} & \multicolumn{2}{c}{Tracking AP} \\
\cmidrule(lr){2-4}\cmidrule(lr){5-6}
        Method & AP\textsuperscript{[0.1,0.8]} & AP\textsuperscript{OoF} & AP & AP & AP\textsuperscript{[0,0.8]} \\
        \midrule
        $\Delta b$ & 13.86 & 14.79 & 28.62 & \textbf{16.47} & 8.94 \\
        $f$ & 16.12 & 17.08 & 29.58 & 16.12 & 10.08 \\
        $f$ and $\Delta b$ & \textbf{16.41} & \textbf{17.64} & \textbf{29.87} & 16.35 & \textbf{10.13} \\
        \bottomrule[1.5pt]
    \end{tabular}
\end{table}


\begin{table}[t]
    \centering
    \caption{\textbf{Number of MLP layers in Amodal Expander.} Empirically, a lightweight 2-layer MLP amodal expander is sufficient to generate reasonable amodal predictions. All models were trained on TAO-Amodal training set for 20k iterations.}
    \label{tab:ablation-MLP-layers}
    \begin{tabular}{lccccc}
        \toprule[1.5pt]
              &   \multicolumn{3}{c}{Detection AP} & \multicolumn{2}{c}{Tracking AP} \\
\cmidrule(lr){2-4}\cmidrule(lr){5-6}
        \# layers & AP\textsuperscript{[0.1,0.8]} & AP\textsuperscript{OoF} & AP & AP & AP\textsuperscript{[0,0.8]} \\
        \midrule
        1-layer & 13.78 & 15.19 & 28.21 & 14.29 & 8.12 \\
        2-layer & \textbf{16.41} & \textbf{17.64} & \textbf{29.87} & \textbf{16.35} & \textbf{10.13} \\
        4-layer & 15.55 & 17.02 & 29.41 & 16.35 & 9.99 \\
        6-layer & 14.55 & 15.64 & 28.79 & 16.05	& 9.09 \\
        \bottomrule[1.5pt]
    \end{tabular}
\end{table}

In this appendix, we extend our discussion of the proposed dataset and method within the context of tracking any object with amodal perception. Specifically, we discuss details about the training and PasteNOcclude technique in Section~\ref{sec:impl_detail}, provide further empirical analysis in Section~\ref{sec:supp-empirical}, and outline the annotation guidelines of our dataset in Section~\ref{sec:ann-guideline}. We further provide comprehensive video demonstration of our dataset and qualitative results at \texttt{webpage/index.html}.

\section{Implementation details}
\label{sec:impl_detail}
\subsubsection{Training Amodal Expander.}
We trained amodal expander on TAO-Amodal training set for 20k iterations for all experiments unless specified. We used a 2-layer MLP as the architecture. The hidden dimension of MLP is 256. We apply ReLU~\cite{relu} and dropout~\cite{srivastava2014dropout} with a probability of 0.2 to each layer except the last one. We implemented the expander in conjunction with GTR~\cite{GTR}. Architecture details of GTR align with the selection in the prior work~\cite{GTR}. We used 0.01 as the base learning rate and applied \texttt{WarmupCosineLR}~\cite{he2019bag} as the scheduler. The optimizer is AdamW~\cite{adamw}. The batch size for training is 4. We adopted the training methodology outlined in~\cite{GTR}, treating each image as an independent sequence. We applied data augmentation~\cite{zhou2020tracking}, including random cropping and resizing, to each image to produce synthetic videos with a length of 8 frames. Beyond this, we further applied PasteNOcclude, introduced in Sec.~\ref{sec:PnO} in the main paper, on top of the synthetic videos to automatically generate more occlusion scenarios. We provide the hyperparameter details of PasteNOcclude in the next section.

\textbf{\\PasteNOcclude (PnO).}
We illustrated visual examples of PnO in Fig.~\ref{fig:PnO}. We mask the background area with the segmentation mask and collect the cropped object from LVIS~\cite{gupta2019lvis} and COCO~\cite{lin2014microsoft} to serve as occluders. We filter out segments where the mask area is less than 70\% of the bounding box area to ensure that the occluder is not occluded. In the training process, we view each image as a sequence and create an 8-frame sequence employing the data augmentation strategy in GTR~\cite{GTR} based on each image. Subsequently, we randomly select 1 to 7 segments from the collection and place them at random locations. Further, we randomly adjust the height and width of the inserted segments within the range of $[12, 192]$. We randomly determine the object's location and size only in the first and last frames to ensure smooth transitions between consecutive frames. The size and location in intermediate frames are obtained through interpolation.

\section{More empirical analysis}
\label{sec:supp-empirical}
We used the evaluation metrics defined in~\cref{sec:benchmark} in the main draft. We summarize all the definitions in Table~\ref{tab:Eval-Metrics}. We presented additional experiments involving state-of-the-art trackers in Section~\ref{sec:supp-off-the-shelf} and the amodal expander in Section~\ref{sec:supp-amodal-expander}.

\subsection{Benchmarking off-the-shelf-trackers}
\subsubsection{Evaluation on TAO-Amodal validation set.}
We report detection and tracking average precision (AP) numbers of SOTA off-the-shelf trackers on TAO-Amodal validation set running at 1fps with an IoU threshold 0.5 in \cref{tab:Off-the-shelf-AP50-1fps}. We also observed similar performance trends when running at 5fps with higher IoU thresholds, shown in~\cref{tab:Off-the-shelf-AP50-5fps,tab:Off-the-shelf-AP-1fps}. Every off-the-shelf tracker was trained on either TAO~\cite{dave2020tao} or LVIS~\cite{gupta2019lvis}, ensuring alignment of category vocabulary with our dataset as detailed in~\cref{sec:benchmark}. We reproduced AOA~\cite{AOA} using their released implementation, with object detector trained on LVIS and tracking ReID head trained on TAO.

\subsubsection{Using off-the-shelf modal trackers for amodal perception.}
Table~\ref{tab:Off-the-shelf-AP50-1fps} reveals notable differences in detection AP between modal (Modal AP) and amodal annotations (AP), amounting to an 8.49\% difference. Additionally, the amodal tracking AP experiences a substantial decline compared to modal tracking AP.
These results highlight the difference between amodal and modal perception.

\subsubsection{How well do standard trackers handle occlusion?}
Existing off-the-shelf trackers exhibit reasonable performance in detecting non-occluded objects, with ViTDet achieving 39.71\% AP\textsuperscript{[0.8,1]} as revealed in Table~\ref{tab:Off-the-shelf-AP50-1fps}. However, all trackers face challenges in handling heavily occluded, partially occluded (AP\textsuperscript{[0.1,0.8]}) and out-of-frame (OoF) scenarios. We noticed that ViTDet operating at 5 fps benefits from the property of SORT to estimate the location in the current frame using past information in~\cref{tab:Off-the-shelf-AP50-5fps}. Nevertheless, this improvement comes at the cost of processing ViT-Det on 5x more frames than models running at 1 fps. In contrast, amodal completion could be a promising way for efficiently handling occlusion.

\begin{table*}[t!]
    \centering
    \caption{\textbf{Evaluation metrics with IoU threshold 0.5.} We define variations of AP~\cite{lin2014microsoft} and Track-AP~\cite{dave2020tao} based on levels of occlusion.}
    \label{tab:Eval-Metrics}
    \resizebox{1.0\textwidth}{!}{
    \begin{tabular}{lll}
        \toprule[1.5pt]
        Metric & \multicolumn{1}{c}{Definition} & Type \\
        \multirow{2}{*}{AP} & Average Precision (AP) averaged across all categories at an & \multirow{7}{*}{Detection Metrics} \\
        & IoU threshold 0.5. & \\
        AP\textsuperscript{[0, 0.1]} & AP for heavily occluded objects, with visibility smaller than 0.1. & \\
        AP\textsuperscript{[0.1, 0.8]} & AP for partially occluded objects, with visibility in [0.1, 0.8]. & \\
        AP\textsuperscript{[0.8, 1.0]} & AP for non-occluded objects, with visibility larger than 0.8. & \\
        AP\textsuperscript{OoF} & AP for partially out-of-frame (OoF) objects. & \\
        Modal AP & AP on modal annotations. & \\
        \midrule
        \multirow{2}{*}{Track-AP~\cite{dave2020tao}} & Average Precision of a track averaged across all categories at an & \multirow{5}{*}{Tracking Metrics}\\
        & 3D IoU threshold 0.5. & \\
        \multirow{2}{*}{Track-AP\textsuperscript{[0, 0.8]}} & Track-AP for any track that is occluded, with visibility & \\
        & at or below 0.8, for more than 5 frames (seconds). & \\
        Modal Track-AP & Track-AP on modal annotations & \\
        \bottomrule[1.5pt]
    \end{tabular}
    }
\end{table*}

\label{sec:supp-off-the-shelf}
\textbf{\\Evaluation with higher IoU thresholds.}
In Table~\ref{tab:Off-the-shelf-AP-1fps}, we evaluate the trackers with average precision (AP) averaged over 10 IoU thresholds from 0.5 to 0.95 at a step 0.05. The performance trend basically aligns with what we observed in Table~\ref{tab:Off-the-shelf-AP50-1fps} in the main paper. GTR~\cite{GTR} obtained strong performance in both detection and tracking. When evaluated with higher IoU thresholds, ViTDet~\cite{ViTDet} and SORT~\cite{SORT} demonstrate inferior detection performance compared to GTR, indicating a contrasting outcome compared to the results obtained at a 0.5 threshold. This shows the limitations of SORT~\cite{SORT} in accurately estimating bounding boxes.

\textbf{\\Running trackers at higher fps.}
We reported the performance of state-of-the-art trackers running at 5 fps in \cref{tab:Off-the-shelf-AP50-5fps}. We noticed that ViTDet~\cite{ViTDet} along with SORT~\cite{SORT} achieved the best performance among all the trackers. This aligns with our intuition as SORT estimates the location in the current frame based on prior-frame locations. This property benefits from running at higher fps, but it requires processing ViTDet on 5\texttimes  more frames than models operating at 1 fps, heavily increasing computational demands.

\subsection{Amodal expander experiments}
\label{sec:supp-amodal-expander}

\subsubsection{Scaling up training data.}
In~\cref{tab:scale-amodal-expander-AP50}, we scale up the training data to 4x by including test videos as train set and evaluate the amodal expander on the validation set. We note that simply increasing the size of the training data does not significantly improve the metrics compared to results shown in~\cref{tab:sota-fine-tune-AP50-1fps}. This validates our design to propose TAO-Amodal as an evaluation benchmark.

\subsubsection{Investigating key information for amodal box inference.}
Table~\ref{tab:ablation-input} reports different input choices to the amodal expander. Modal box (deltas) $\Delta b$, output by the regression head as shown in Fig.~\ref{fig:Amodal-Expander} in the main paper, are used to yield final modal box predictions when applied to region proposals and thus contain information about the exact location of modal box predictions. Proposal features includes visual appearance information of the detected region proposals. Absence of visual cues significantly diminishes the performance of both detection and tracking under occlusion. Interestingly, the amodal expander, incorporating both modal delta and proposal features, yielded the most favorable outcomes. This indicates that estimating modal box locations also contributes to effective amodal reasoning.

\subsubsection{Number of MLP layers.}
We tested with the depth of amodal expander architecture in Table~\ref{tab:ablation-MLP-layers}. We observe a reverse-U pattern concerning the number of MLP layers, with two-layer MLPs demonstrating superior performance compared to other models. A one-layer MLP proves suboptimal in both detection and tracking. Notably, using a 1-layer MLP results in slightly inferior outcomes compared to fine-tuning the regression head, as indicated in Table~\ref{tab:amodal-expander-AP50} in the main paper. We argue that the regression head may derive benefits from pre-training on modal benchmarks.

\begin{table*}[t!]
    \centering
    \caption{\textbf{Annotation guidelines.} TAO-Amodal is annotated with the guidelines above, which taxonomizes occlusions across severity (partial versus complete) and type (in/out-of-frame). As mentioned in Sec.~\ref{sec:annotation} in the main paper, we scope out the case where an object may be present behind the camera. For out-of-frame occlusions, we limit the \textit{annotation workspace} to be twice the image size.}
    \label{tab:guidelines}
    \resizebox{1.0\textwidth}{!}{
    \begin{tabular}{llll}
        \toprule[1.5pt]
        Occlusion type & Extent & Cases & Instructions \\
        \midrule
        \multirow{7}{*}{In-frame} & Partial & Partially occluded before being fully visible & Annotate with best estimate using category label \\
        & & Partially occluded after being fully visible & Annotate with best estimate \\
            & Complete &  Invisible before being (partially) visible & Only annotate if the object has been visible before \\
            & & Invisible after being (partially) visible & If confident, annotate with best estimate \\
            & & & If not, only annotate till the last visible frame \\
            & & Invisible for a while & If confident, annotate with best estimate \\
            & & & If not, still annotate but add an uncertainty flag \\
        \midrule
        \multirow{3}{*}{Out-of-frame} & Partial & Object goes beyond image border & Only annotate inside the annotation workspace \\
            & & Object goes beyond the padded image & Clip at the border of the padded image \\
            & Complete & - & -\\
        \midrule
        \multirow{2}{*}{Behind-the-frame} & Partial & Object is in front of and behind the camera & Only label the part of object in front of camera \\
            & Complete & - & - \\
        \bottomrule[1.5pt]
    \end{tabular}
    }
\end{table*}

\section{Annotation guidelines}
\label{sec:ann-guideline}
We ensure high-quality annotations by requiring annotators to follow the guidelines detailed in Table~\ref{tab:guidelines}. Our coverage spans various occlusion scenarios, encompassing in-frame, out-of-frame, or behind-the-scene situations, where an object may be partially obscured behind the camera.

\clearpage

\end{document}